\documentclass[10pt,twocolumn,letterpaper]{article}

\usepackage[pagenumbers]{cvpr} %
\usepackage[accsupp]{axessibility}  %

\usepackage{graphicx}
\usepackage{amsmath}
\usepackage{amssymb}
\usepackage{booktabs}

\usepackage{multicol}
\usepackage{multirow}
\usepackage{soul}
\usepackage[pagebackref,breaklinks,colorlinks]{hyperref}

\usepackage[capitalize]{cleveref}
\crefname{section}{Sec.}{Secs.}
\Crefname{section}{Section}{Sections}
\Crefname{table}{Table}{Tables}
\crefname{table}{Tab.}{Tabs.}

\def\cvprPaperID{4910} %
\def\confName{CVPR}
\def\confYear{2023}

\begin{document}

\title{AnyFlow: Arbitrary Scale Optical Flow with Implicit Neural Representation}

\author{Hyunyoung Jung$^1$\thanks{This work was done during the author's internship at Meta.} \qquad
Zhuo Hui$^2$
\qquad
Lei Luo$^2$
\qquad
Haitao Yang$^2$
\\
Feng Liu$^2$\thanks{Affiliated with Meta at the time of this work.}
\qquad
Sungjoo Yoo$^1$
\qquad
Rakesh Ranjan$^2$
\qquad
Denis Demandolx$^2$ 
\vspace{-1mm}
\\\\
$^1$Seoul National University \qquad $^2$Meta Reality Labs
}

\maketitle

\begin{abstract}

To apply optical flow in practice, it is often necessary to resize the input to smaller dimensions in order to reduce computational costs. However, downsizing inputs makes the estimation more challenging because objects and motion ranges become smaller. Even though recent approaches have demonstrated high-quality flow estimation, they tend to fail to accurately model small objects and precise boundaries when the input resolution is lowered, restricting their applicability to high-resolution inputs. In this paper, we introduce AnyFlow, a robust network that estimates accurate flow from images of various resolutions. By representing optical flow as a continuous coordinate-based representation, AnyFlow generates outputs at arbitrary scales from low-resolution inputs, demonstrating superior performance over prior works in capturing tiny objects with detail preservation on a wide range of scenes. We establish a new state-of-the-art performance of cross-dataset generalization on the KITTI dataset, while achieving comparable accuracy on the online benchmarks to other SOTA methods.

\end{abstract}

\section{Introduction}

Optical flow seeks to estimate per-pixel correspondences, characterized as the horizontal and vertical shift, from a pair of images. Specifically, it aims to identify the correspondence across pixels in different images, which is at the heart of numerous computer vision tasks such as video denoising~\cite{yu2020joint, buades2016patch, liu2010high}, action recognition~\cite{sun2018optical, wang2013action, wang2019hallucinating} and object tracking~\cite{decarlo2000optical, kale2015moving, schwarz2012human}. This is particularly challenging due to the fact that the scene geometry and object motion are combined into a single observation, making the inverse problem of estimating motion highly ill-posed.

A common assumption for enabling computationally tractable optical flow is to explicitly account for small motion in local neighborhood and incorporate additional prior to constrain the solution space, such as using total variation prior~\cite{tvl1, 6082986} and smooth prior~\cite{Kanade, 378214, qt}. While effective, these techniques are significantly restricted by the assumption and do not generalize well for real-life scenes with sophisticated geometry and motions.

\begin{figure}[t!]
    \centering
    \includegraphics[width=0.975\linewidth]{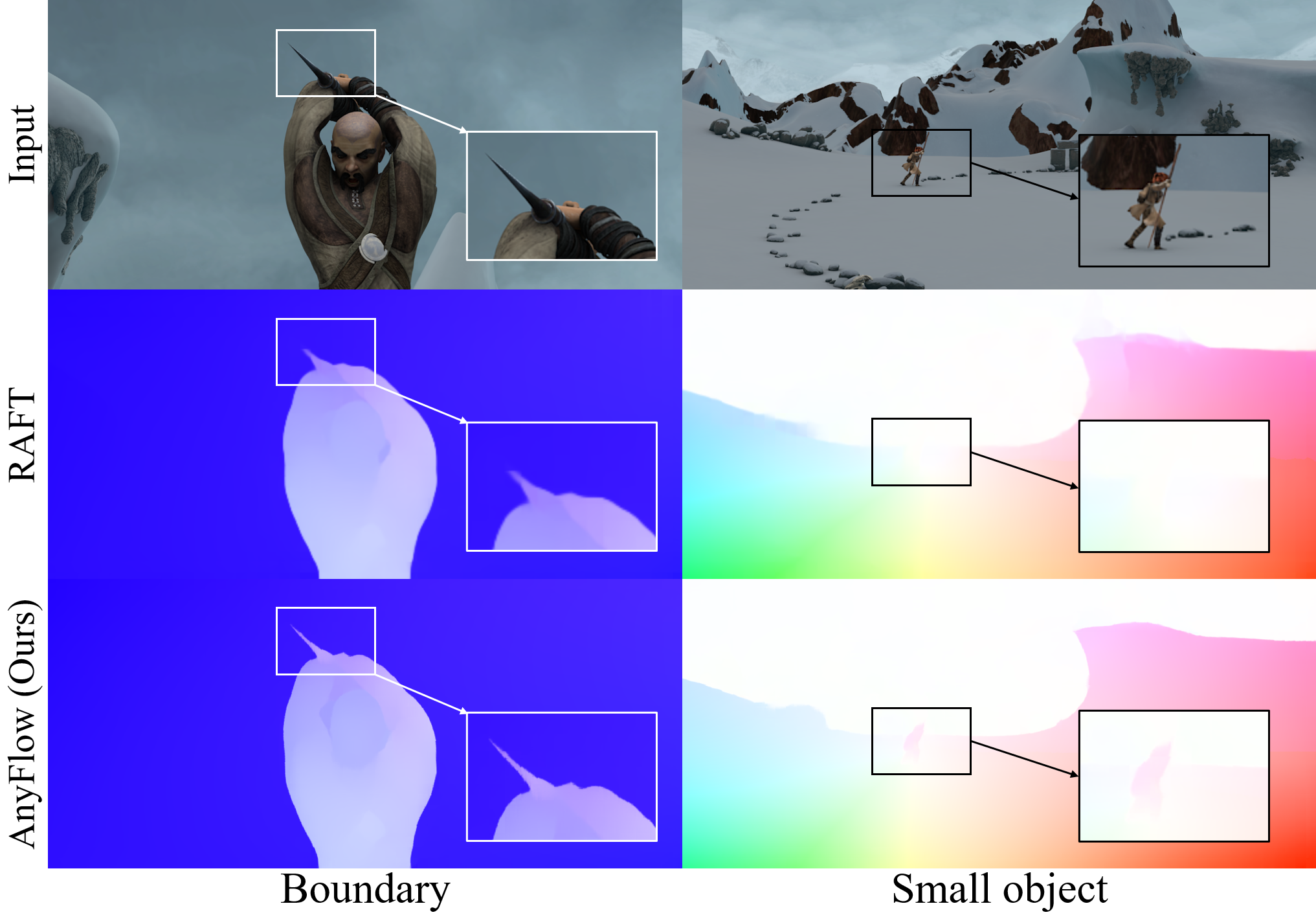}
    \caption{Predicting $50\%$ downsized images on Sintel~\cite{sintel}, we compare AnyFlow to RAFT~\cite{raft}. AnyFlow shows clearer boundaries, accurate shapes, and better detection of small objects.
    }
    \label{fig:intro}
\end{figure}
An alternative approach is motivated by the success of deep neural networks. Different from using handcrafted priors, learning-based methods~\cite{flownet,  yang2019vcn, zhao2020maskflownet} design end-to-end networks to regress the motion field. Sun et al.~\cite{pwcnet} use the coarse-to-fine approach to characterize the pixel-to-pixel mapping and several methods~\cite{hui2018liteflownet, ranjan2017optical, yang2019vcn, jahedi2022multi} are developed along this line. The drawback of these approaches is that the accuracy of flow estimation is not only limited by the sampling but also has a critical dependence on a good initial solution because the pixel correspondence is fundamentally ambiguous. Tedd and Deng~\cite{raft} resort to iterative update module on top of the correlation volumes, allowing better estimates in tiny and fast-moving objects. Indeed the success of this model relies on having a correlation volume that is high quality in characterizing the feature correspondence for sub-pixel level. Inspired by its success, follow-up methods~\cite{ jiang2021learning, Luo_2022_CVPR, jeong2022imposing, sun2022disentangling, sun2021autoflow} have further improved it in multiple ways. One popular direction is to introduce attention mechanisms~\cite{vaswani2017attention} for efficiency~\cite{flow1d, separableflow}, occlusion handling~\cite{gma}, and large displacements~\cite{craft}. As attention has been found to be effective in estimating long-range dependencies, Xu et al.~\cite{xu2022gmflow} and Huang et al.~\cite{huang2022flowformer} adopt Vision Transformers~\cite{liu2021swin} to perform global matching and demonstrate their effectiveness. However, these methods naturally become less effective for low-resolution images, where inaccuracy is introduced when computing the correlation volumes and the iterative refinements. They tend to fail in accurately modeling small objects and precise boundary when the input resolution is lowered. This inherently limits the applicability for mobile devices in particular, where resizing the input to small size is often necessary to reduce the computational cost.

In this paper, we propose AnyFlow, a novel method that is agnostic to the resolution of images. The key insight behind AnyFlow is the use of implicit neural representation (INR) for flow estimation. INRs, such as LIIF~\cite{liif}, have been demonstrated to be effective for image super-resolution, as they model an image as a continuous function without limiting the output to a fixed resolution. This enables image generation at arbitrary scales. Inspired by this capability, we design a continuous coordinate-based flow upsampler to infer flow at any desired scale.
In addition, we propose a novel warping scheme utilizing multi-scale feature maps together with dynamic correlation lookup to generalize well on diverse shapes of input.
As the proposed methods produce a synergistic effect, we demonstrate superior performance for tiny objects and detail preservation on a wide range of scenes involving complex geometry and motions. Specifically, we verify the robustness for the input with low resolution. Fig.~\ref{fig:intro} showcases our technique against RAFT~\cite{raft}. Our contributions are summarized as follows:
\begin{itemize}
    \item 
    We introduce AnyFlow, a novel network to produce high quality flow estimation for arbitrary size of image.
    \item 
    We present the methods to utilize multi-scale feature maps and search optimal correspondence within predicted range, enabling robust estimation in wide ranges of motion types.
    \item 
    We demonstrate strong performance on cross-dataset generalization by achieving more than 25\% of error reduction with only 0.1M additional parameters from the baseline and rank 1st on the KITTI dataset.
\end{itemize}
Together, our contributions provide, for the first time, an approach for \textit{arbitrary} size of input, which significantly improves the quality for low-resolution images and extends the applicability for portable devices.

\label{sec:intro}

\section{Related Work}

\paragraph{Physical based flow method.}
Estimating motion from images has been a long-standing goal in computer vision. Under the common assumption that the brightness across consecutive images remains constant, it has been addressed by utilizing a smooth prior~\cite{Kanade, 378214, qt, horn1981determining} and noise estimates by incorporating variational prior~\cite{tvl1, 6082986}. However, the induced physical-based priors are inherently limited in their ability to provide precise approximation to the real-world.

\paragraph{Learning based flow method.}
Deep neural networks have shown unprecedented power in solving ill-posed problems.
In the context of optical flow, Dosovitskiy et al.~\cite{flownet} first train a network end-to-end,
utilizing correlations between deep feature vectors. To capture large motions, coarse-to-fine methods~\cite{pwcnet,hui2018liteflownet, hofinger2020improving, zhao2020maskflownet, yin2019hierarchical} estimate flow from low resolution and refine at high resolution. Stacked networks in series with warping~\cite{ilg2017flownet} and pyramidal feature networks~\cite{yang2019vcn, ranjan2017optical} are adopted to refine flow field. Since the initial flow is estimated at the coarse stage with low-resolution, they mainly suffer from difficulties in detecting small and fast-moving objects.
Recently, Tedd and Deng~\cite{raft} extend the idea by first introducing the recurrent layer to refine a single flow field, obviating the need for multi-resolution search on feature matching.  
Based on its success, further improvements were made by leveraging patch-wise estimation~\cite{zheng2022dip, Luo_2022_CVPR}, sparsity~\cite{jiang2021learning}, graph reasoning~\cite{luo2022learning} and development of training strategies~\cite{jeong2022imposing, sun2022disentangling, sun2021autoflow}.  
One of the recent popular directions is to utilize attention mechanism~\cite{vaswani2017attention} to improve efficiency~\cite{flow1d, separableflow}, resolve occlusion~\cite{gma}, and capture large displacements~\cite{craft}. 
Since self-attention proves useful in estimating long-range dependency, vision transformer~\cite{liu2021swin} was adopted for optical flow~\cite{xu2022gmflow, huang2022flowformer} to perform global matching and demonstrated effectiveness. 
While these methods enable high-quality estimation, the prediction for tiny objects is still a challenge due to the fact that their underlying architectures estimate the flow in fixed accuracy, making it sensitive to resolution changes.
In contrast, our method processes images with arbitrary size without affecting the quality in flow estimation.

\begin{figure*}[t!]
    \centering
    \includegraphics[width=0.95\linewidth]{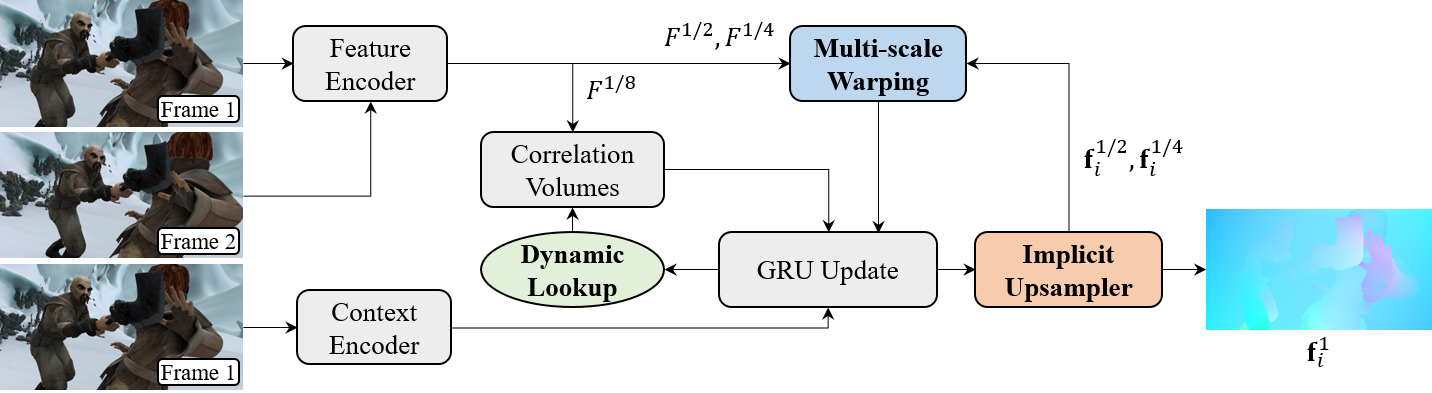}
    \caption{{\bf Overall architecture of AnyFlow.} 
    AnyFlow is built upon the RAFT~\cite{raft} framework: feature extraction, all-pairs correlation volumes, and GRU update, as depicted by the gray blocks. After building the correlation volumes, the network performs a dynamic lookup for sampling values within an input-dependent local grid. The neural implicit flow upsampler rescales the output flow to arbitrary scales and generates multi-scale optical flows ($\mathbf{f}_i^{1/2}, \mathbf{f}_i^{1/4}$) for the multi-scale feature warping module, which in turn warps feature maps from the feature encoder. The GRU Update combines these elements with its output feature $M$ to produce a residual flow $\Delta\mathbf{f} \in \mathbb{R}^{H/8\times W/8 \times 2}$.
    }
    \label{fig:overall}
\end{figure*}

\paragraph{High-resolution flow from low-resolution image.}
High-resolution (HR) optical flow from low-resolution (LR) images is desired for various tasks. For example, in video super-resolution, temporal consistency of HR frames is crucial and optical flow~\cite{tu2022optical, caballero2017real, tao2017detail} is required to achieve it. 
Sajjadi et al.~\cite{sajjadi2018frame} use a recurrent approach to reconstruct HR frame, while maintaining temporal consistency by bilinearly interpolating LR flow. Wang et al.~\cite{wang2020deep} propose a coarse-to-fine network with motion compensation to infer HR flows from LR images. However, the interpolation incurs artifacts and CNN-based coarse-to-fine approach limits the resolution to fixed scale. In addition, such task requires precise flow estimation from LR images, since inaccurate motion leads to severe distortion in HR reconstruction. 
We demonstrate robust estimation from LR images and resize flow into arbitrary resolution, extending our applicability to such downstream task.

\paragraph{Implicit neural representation.}
Implicit neural representation (INR), parameterized as an MLP, maps a spatial coordinate to output signal. It has been widely adopted to represent learning-based 3D geometry~\cite{nerf, pumarola2020d,wang2021nerfmm,kaizhang2020,chen2021mvsnerf,pixelnerf,park2021nerfies,lin2021barf,barron2021mipnerf} and several studies explore it in 2D tasks as well such as image and shape generation~\cite{Karras2021, implicit1}, image fitting~\cite{sitzmann2019siren} and semantic segmentation~\cite{ifa}. Recently, for super-resolution~\cite{videoinr, liif, ultrasr}, Chen et al.~\cite{liif} propose LIIF that learns continuous image representation from local latent code. Different from the previous methods~\cite{dong2015image, lai2017deep, lim2017enhanced, ledig2017photo} that upsample image into a fixed set of scales, LIIF models an image as a continuous function and the output is not limited to the fixed resolution, enabling image generation at arbitrary scales. Based on it, Chen et al.~\cite{videoinr} design an implicit network to learn intermediate motion fields and maintain temporal consistency for continuous video super-resolution. However, the motion network does not perform precise estimation and has difficulty handling large motion.
Different from them, we re-design LIIF as a novel flow upsampler and integrate it into the existing optical flow framework~\cite{raft}, thereby enabling us to generate optical flow in arbitrary scales with robust estimation in diverse scenarios.

\section{Arbitrary Scale Optical Flow}

\paragraph{Problem statement.}
Given a pair of input images, $I_1$ and $I_2\in\mathbb{R}^{H\times W \times 3}$, optical flow seeks to solve per-pixel motion fields
$\mathbf{f}\in\mathbb{R}^{H\times W \times 2}$, which account for horizontal and vertical motion. Our goal is to estimate the flow as an implicit function of an input image pair for arbitrary scales as
\[\mathbf{f}^s = \mathcal{I}(I_1, I_2), \]
where $\mathbf{f}^s\in\mathbb{R}^{sH\times sW \times 2}$ for scale factor $s$, and $\mathcal{I}$ denotes the underlying function we expect to model via the network.

\paragraph{Overview.}
We design AnyFlow based on the general stages of RAFT~\cite{raft}. We first extract multi-scale feature maps through the feature encoder, compute correlation volumes and update the flow field by accumulating the residual flow. Built upon it, we propose Neural Implicit Flow Upsampler (Sec.~\ref{sec:liif}) to rescale the accumulated flow into arbitrary scale, including restoring the original resolution from downscaled inputs. Our proposed Multi-scale Feature Warping module (Sec.~\ref{sec:HR}) enables utilization of high-resolution representations for flow prediction. Finally, we propose Dynamic Lookup (Sec.~\ref{sec:dynamic}) to search correspondence from the correlation volumes depending on input images. The overall architecture is illustrated in Fig.~\ref{fig:overall} and we describe the methods in detail in the following sections.

\subsection{Neural Implicit Flow Upsampler}
\label{sec:liif}
\paragraph{Iterative refinement.}
Following RAFT, the network first produces $1/8$-sized residual flow $\Delta\mathbf{f} \in \mathbb{R}^{H/8\times W/8 \times 2}$ and accumulates it by GRU iterations. After $i^{th}$ update, the accumulated flow $\mathbf{f}_{i} \in \mathbb{R}^{H/8\times W/8 \times 2}$ is computed as: $\mathbf{f}_{i} = \Delta\mathbf{f}_i + \mathbf{f}_{i-1}$. 
Note that, in the following sections, the flow {\it without superscript}, i.e., $\mathbf{f}_{i}$, always denotes the accumulated flow in $1/8$ size of the input resolution .
\paragraph{Local implicit image function.} 
We utilize LIIF~\cite{liif} to generate an arbitrary scale of output from a fixed input signal. In the LIIF representation, output signal $o$ at 2D continuous coordinate $x=(u,v)$ is estimated from a decoding function $f_\theta$, parameterized as an MLP as: $o = f_\theta(z, x)$. Here, $z$ is a feature vector sampled from the 2D feature map $M$ at location $x$, and $M$ is extracted from input signal $I$ through encoding function $\mathcal{E}$. That is, the decoding function $f_\theta$ maps each coordinate $x$ to the output signal, depending on the feature encoding $z$. 

\paragraph{Upsample in arbitrary scale.} 
We first extract 2D feature map $M\in \mathbb{R}^{H/8\times W/8 \times C}$ from the input pair,  $I_1$ and $I_2$. 
\begin{align}
    & M = \mathcal{E}(I_1, I_2)
    \label{eq:encoding}
\end{align}

The encoding function $\mathcal{E}$ is composed of modules ranging from the feature encoder to the GRU, as described in Fig.~\ref{fig:overall}, with $M$ representing the GRU's hidden state.

Similar to RAFT, we represent the motion of each pixel in the upsampled resolution as a convex combination of $3\times3$ local neighbors in the coarse flow, $\mathbf{f}_{i} \in \mathbb{R}^{H/8\times W/8 \times 2}$. The convex weights, referred to as $mask$, are symbolized by $\mathcal{O}$ and defined at a continuous coordinate $x_q$ as:

\begin{align}
    & \mathcal{O}(x_q)=f_\theta(z^*, x_q-v^*, \psi(x_q-v^*))
    \label{eq:mask}
\end{align}

\noindent{}, where $z^*$ represents the nearest feature vector from $x_q$ in $M$, and $v^*$ corresponds to the coordinate of $z^*$.
To learn high-frequency signals, we apply positional encoding~\cite{ifa, ultrasr} $\psi$ to the relative position, $x_q-v^*$. The inputs are then fed into the decoding function $f_\theta$, parameterized as an MLP.

The output, $\mathcal{O}(x_q)$, represents a $3\times3 \times n^2$-dimensional vector where the first two dimensions ($3\times3$) correspond to the convex weights of $3\times 3$ neighbors in the coarse resolution. $n$ is a fixed hyperparameter, indicating that we generate an $n \times n$ local patch in the higher resolution with a single query at $x_q$.

To upsample $\mathbf{f}_{i}$ into our desired resolution, $H_o \times W_o$, we first evenly sample the query points $x_q$ from a continuous 2D grid. The horizontal and vertical ranges of the grid are within the size of $M$. Since a single query upsamples a single pixel of flow to an $n\times n$ patch, we sample $\frac{H_o}{n} \cdot \frac{W_o}{n}$ numbers of the query points and feed them into $f_\theta$ to produce $\mathcal{O}$, which upsample $\mathbf{f}_{i}$ into $H_o \times W_o$ resolution.

Unlike RAFT, which relies on a fixed shape of the mask to achieve only $8-$times upsampling, we offer greater flexibility by allowing for general resizing through the adjustment of the number of samples.

\begin{figure}[t!]
    \centering
    \includegraphics[width=0.975\columnwidth]{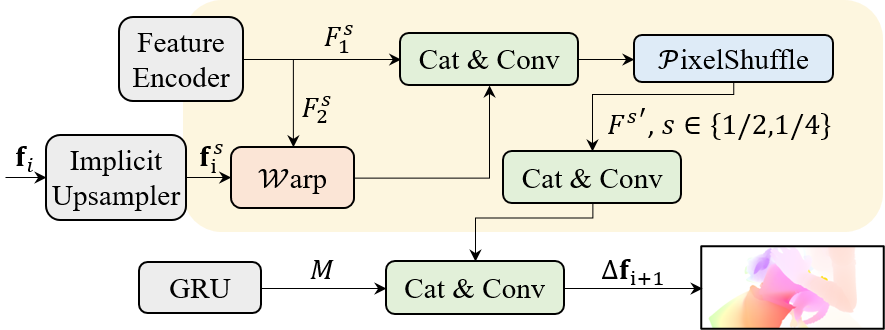}
    \caption{{\bf Illustration of the multi-scale feature warping.} To utilize high-resolution representation, multi-scale feature maps, ${F^s}$, $s\in \{1/2, 1/4\}$, are combined with GRU feature, $M$. The combined features are used to predict the residual flow, $\Delta\mathbf{f}$. }
    \label{fig:HR}
\end{figure}

\subsection{Multi-scale Feature Warping}
\label{sec:HR}
RAFT's capacity to capture small objects is limited by its reliance on ${1/8}$-scaled intermediate features. Moreover, the intricacy of 4D correlation volumes hinders its capacity to benefit from high-resolution features. To address this, we compute a partial cost volume through warping to effectively utilize high-resolution feature maps.

Following feature extraction, we obtain multi-scale feature maps ${F^s}$, where $s \in \{1/2, 1/4\}$, from each input frame, as well as a pair of $F^{1/8}$ maps for calculating correlation volumes. Given a pair of feature maps, $F^s_1$ and $F^s_2$, we inversely warp $F^s_2$ using the upsampled flow prediction, $\mathbf{f}^s_{i}$, which is derived by re-scaling the accumulated flow $\mathbf{f}_{i}$ with the implicit upsampler. We then concatenate $F^s_1$ and the warped one $\mathcal{W}(F^s_2, \mathbf{f}^s_i)$, followed by applying a $1 \times 1$ convolution to reduce the channel dimension.

Note that we directly upsample intermediate flow by reusing the implicit upsampler, removing the need for pyramidal architectures~\cite{pwcnet, yang2019vcn, hui2018liteflownet, ranjan2017optical, jahedi2022multi} and extra parameters.

\paragraph{Downsample by PixelShuffle.}

Our objective is to integrate the outputs from the prior step with the GRU feature $M$ and employ them to estimate the residual flow. First, we must ensure that the spatial dimensions align, given that the outputs possess spatial dimensions of $sH \times sW$, while $M$ is at a $1/8$ scale. To achieve this, we apply the PixelShuffle~\cite{pixelshuffle} (denoted by $\mathcal{P}$) to downscale the dimensions to match those of $M$ and concatenate the local neighborhoods along the channel dimensions. This approach allows us to fully exploit spatial information while simultaneously lowering the resolution.

Following this, we apply an additional convolution to the PixelShuffle outputs, ${F^{1/2}}^\prime$ and ${F^{1/4}}^\prime$, and use them to estimate the residual flow for the subsequent iteration, $\Delta \mathbf{f}_{i+1}$, in conjunction with $M$. We outline the entire process in Eqns.~\ref{eq:PixelShuffle}-\ref{eq:Conv} and Fig.~\ref{fig:HR}.

\begin{align}
    \label{eq:PixelShuffle}
    & \Delta \mathbf{f}_{i+1} = \textrm{Conv}([M, \textrm{Conv$_{1\times 1}$}([{F^{1/2}}^\prime, {F^{1/4}}^\prime])]), \\
    & \text{where} \; {F^{1/2}}^\prime = \mathcal{P}(\textrm{Conv$_{1\times 1}$}(F_1^{1/2}, \mathcal{W}(F^{1/2}_2, \mathbf{f}^{1/2}_i))), \\
    & \text{and} \; {{F^{1/4}}}^\prime = \mathcal{P}(\textrm{Conv$_{1\times 1}$}(F_1^{1/4}, \mathcal{W}(F^{1/4}_2, \mathbf{f}^{1/4}_i)))
    \label{eq:Conv}
\end{align}

\begin{figure}[t!]
    \centering
    \includegraphics[width=0.85\columnwidth]{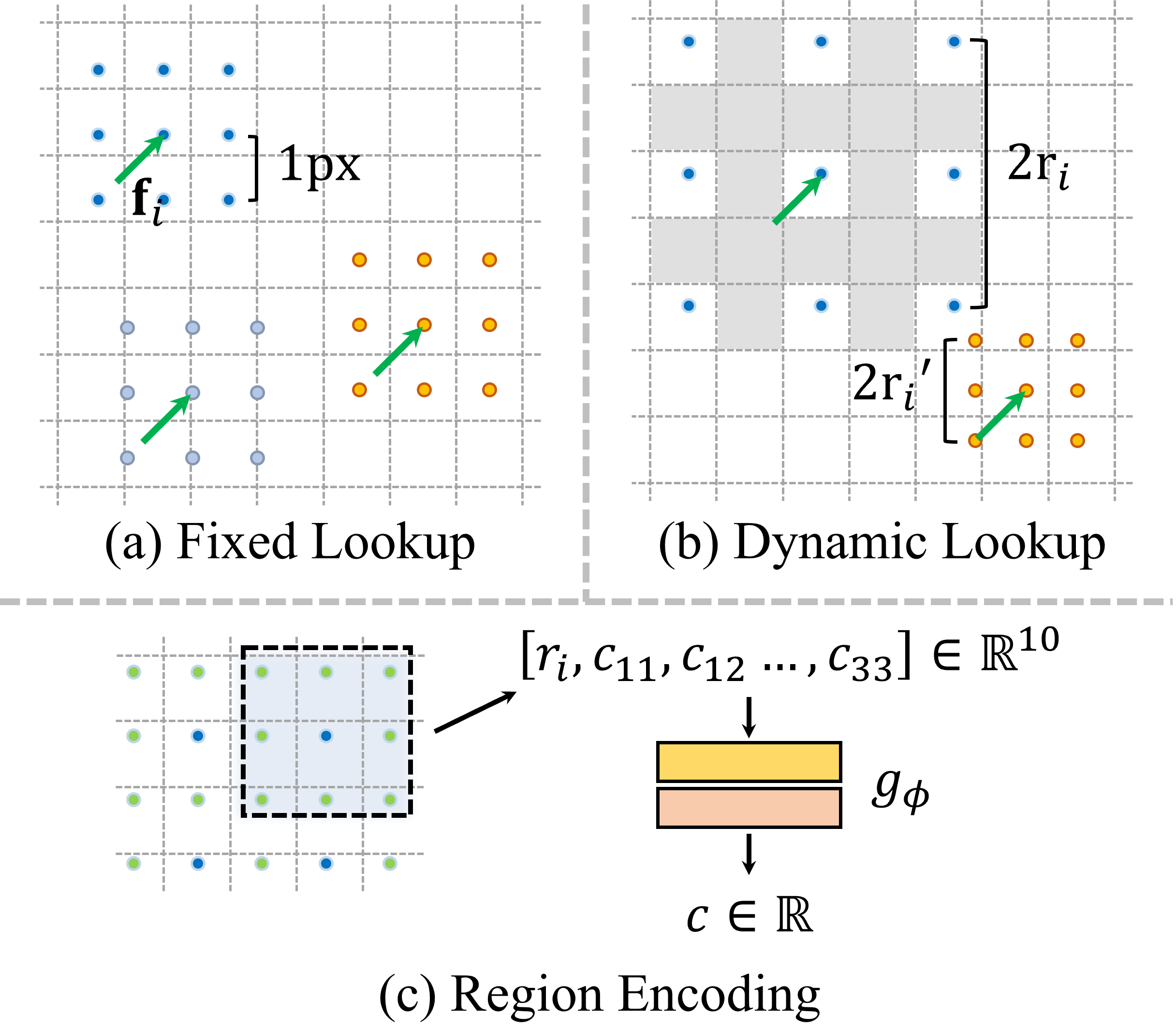}
    \caption{{\bf Comparisons of lookup strategies.} (a) Fixed lookup in RAFT defines local grid in a fixed size and each grid point has 1px apart. (b) Dynamic lookup controls size of the local grid while using the same number of samples. When the grid size increases, there exist blindspots (gray area) between samples. (c) Auxiliary points (green) are defined between each real point (blue). $3 \times 3$ auxiliary region where each real point is the center are fed into the MLP ($g_\theta$) to encode regional correlation.}
    \label{fig:dynamic}
\end{figure}

\subsection{Dynamic Lookup with Region Encoding}
\label{sec:dynamic}
\paragraph{Fixed lookup in RAFT.}
RAFT first maps each pixel $\mathbf{x}$ in $I_1$ to its estimated correspondence $\mathbf{x}'$ in $I_2$ with the current flow $\mathbf{f}_{i}$, as $\mathbf{x}'=\mathbf{x} + \mathbf{f}_i$. It defines a local grid around $\mathbf{x}'$ as {\it the set of integer offsets} within a radius $r$ and bilinearly samples the values of the local grid from each grid location of correlation volume. The radius $r$ is set as a hyperparameter; hence, the single $r$ is applied to every pixel in every GRU iteration. Here, the length of each side of the local grid is defined as $2r$ and the number of samples in each grid is $(2r+1)^2$. 
The samples from each local grid are concatenated along the channel dimension and fed into the GRU.
We call this way {\it Fixed Lookup}, illustrated in Fig.~\ref{fig:dynamic} (a).\footnote{For simplicity, in the figure, we assume $3 \times 3$ local grid.}

\paragraph{Dynamic lookup.}
We can benefit from various sizes of the local grid. When $r$ increases, large displacements can be captured well within each local grid. Small $r$ is beneficial for capturing small motions accurately by concentrating on a small area. To generalize well on diverse shapes of inputs and ranges of motions, we dynamically predict the lookup range and update it for the correlation sampling. 

To this end, the network predicts $r$ as the pixel-wise output while keeping the number of grid points without change, which is updated by the GRU iteration. To be specific, the network predicts residual radius $\Delta r$, and accumulates it as $r_i = \Delta r_i + r_{i-1}$ in the same manner as the flow updates. We use the same number of samples in each grid as in the fixed lookup and the initial radius $r_0$ is set as a hyperparameter.

\paragraph{Region encoding.} 
As $r_i$ is updated, the length of the local grid can exceed the sample number in each side, making the sample interval larger than $1$px. In such cases, sampling on the fixed number of points incurs blindspots (gray regions in Fig.~\ref{fig:dynamic} (b)), where the correlation values are not sampled to compute the output. To alleviate this problem, we introduce region encoding. 
We first define auxiliary points (green points in Fig.~\ref{fig:dynamic} (c)) between the existing points of each grid. We set a $3 \times 3$ auxiliary local region on each existing grid location, where the original grid location (blue points in Fig.~\ref{fig:dynamic} (c)) becomes the center of the auxiliary local region. 

We define the encoding function $g_\phi$, parameterized as an MLP, to learn the mapping from each auxiliary local region to a single correlation value. We also feed $r_i$ as additional input to make the output depend on the area of respective region. 
After obtaining the correlation values with the region encoding, the values of each local grid are then concatenated along the channel dimension.
In this way, we can utilize more sample points and encode regional information to remove the blindspots, without increasing the dimension of the correlation feature maps.

\subsection{Multi-scale Training.}
During training, we randomly downsample input RGB pairs with a probability of $p$. The scale factor is sampled from a uniform distribution for height and width, respectively. AnyFlow takes the downsampled input and generates output in the original size. We supervise it with RAFT loss~\cite{raft}, which minimizes the L1 distance between ground-truth and the predictions from multiple iterations. 
\begin{align}
    & \mathcal{L} = \sum^N_{i=1}{\gamma^{N-i}\lVert \mathbf{f}_{gt} - \mathbf{f}_i^1 \rVert_1}
\end{align}

\begin{table*}[t!]
  \centering
  \begin{tabular}{@{}c c c c c  c c c c@{}}
    \toprule
     Training & \multicolumn{4}{c}{C + T} & \multicolumn{3}{c}{C + T + S + K + H} \\
     \midrule
     
    \multirow{2}{*}{Method} & \multicolumn{2}{c}{Sintel (train)} & \multicolumn{2}{c}{KITTI (train)} & \multicolumn{2}{c} {Sintel (test)} & KITTI (test) & Params  \\
    \cmidrule(lr){2-3} \cmidrule(lr){4-5} \cmidrule(lr){6-7}\cmidrule(lr){8-8}
      &  Clean & Final & F1-epe & F1-all & Clean & Final & F1-all & (M) \\
    \midrule
     PWC-Net~\cite{pwcnet}                  &    2.55    &  3.93     &    10.35    &  33.7       &  - &  - &  - & 9.4       \\
     RAFT~\cite{raft}                       & 1.43  & 2.71  & 5.04  & 17.4   & 1.61 & 2.86 & 5.10  & 5.3       \\
     Flow1D~\cite{flow1d}                   &  1.98     &  3.27      &    6.69    &  22.95       &2.24    & 3.81  & 6.27  & 5.7       \\
     SCV~\cite{jiang2021learning}           & 1.29  & 2.95  & 6.80  &  19.3  &  1.72  &3.60    &6.17   & 5.3           \\
     GMA~\cite{gma}                         & 1.30    & 2.74  & 4.69  & 17.1   & 1.39  & 2.47  & 5.15  &       5.9 \\
     Separable Flow~\cite{separableflow}    & 1.30    &  2.59  & 4.60  & 15.9  &   1.50   & 2.67    & \underline{4.64}   &     6.0  \\
     AGFlow~\cite{luo2022learning}          & 1.31    &  2.69  & 4.82  & 17.0  &     1.43   & 2.47  & 4.89   &     5.6 \\
     DIP~\cite{zheng2022dip}                & 1.30       & 2.82        & 4.29       & \underline{13.73} &  1.44   & 2.83  & {\bf 4.21}   & 5.4 \\
     CRAFT~\cite{craft}                     & 1.27    & 2.79   & 4.88  & 17.5  &  1.45 &  2.42 & 4.79 &   6.3    \\
     GMFlowNet~\cite{zhao2022global}        & 1.14    & 2.71   & 4.24  & 15.4  &  1.39  &  2.65   &  4.79  &   9.3    \\
     GMFLow~\cite{xu2022gmflow}             & 1.08 & 2.48  & 7.77  & 23.4  & 1.74 & 2.90 & - &  4.7     \\
     SKFLow~\cite{sun2022skflow}            & 1.22&  \underline{2.46} &  4.27 &  15.5 &1.28  & \underline{2.23} & 4.84&   6.3    \\
     FlowFormer (small)~\cite{huang2022flowformer} &  1.20 & 2.64 & 4.57 & 16.62   &   -  & -  & -  & 6.2 \\

     FlowFormer~\cite{huang2022flowformer}  & {\bf 0.64} & {\bf 1.5}  & \underline{4.09}  & 14.72  &   {\bf 1.16}    & {\bf 2.09}  & 4.68  & 18.2      \\
     \cmidrule(lr){1-9}
     {\bf AnyFlow ({\it dynamic}) }   &  1.17 & 2.58 & {\bf 3.95} & {\bf 13.01} & \underline{1.23}  & 2.44  & \underline{4.41}  & 5.4   \\
     {\bf AnyFlow ({\it R.E.})}   & 1.10 / \underline{1.05} & 2.52 & {\bf 3.76}  & {\bf 12.44}  & \underline{1.26} & 2.63  &   \underline{4.64} & 5.4   \\
     {\bf AnyFlow + GMA~\cite{gma}}   & 1.16  & 2.62  &  {\bf 4.05} & 13.74 & \underline{1.21} & 2.46 & - & 6.0   \\

    \bottomrule
  \end{tabular}
  \caption{{\bf Quantitative comparisons with recent state-of-the-arts.} The left half results (C+T) are for cross-dataset generalization. All methods are evaluated on train set of Sintel and KITTI dataset after they are trained on FlyingChairs and FlyingThings (C+T). The right half results are fine-tuned on the target datasets and evaluated on the online benchmark. {\it dynamic} means the proposed dynamic lookup and {\it R.E.} denotes the region encoding. We describe the best results in {\bf bold} and the second best results in \underline{underline}.}
  \label{table:comparison}
\end{table*}
\section{Experiments}
Following the previous works~\cite{raft, gma, craft, xu2022gmflow}, we first pretrain AnyFlow on the synthetic datasets. We train for 120K iterations on FlyingChairs~\cite{flownet} and for another 240k iterations on FlyingThings~\cite{things} with batch size of 16. We evaluate the pretrained model on the training set of Sintel~\cite{sintel} and KITTI-2015~\cite{kitti} to compare cross-dataset generalization performance. Subsequently, we perform fine-tuning on the training set of Sintel combined with FlyingThings, KITTI, and HD1K~\cite{hd1k} datasets for 120k iterations with batch size of 8. Then, the model is evaluated on the Sintel online benchmark. We apply warm start strategy~\cite{raft, gma, craft} for it. Finally, we perform additional fine-tuning on the KITTI for 100k iterations and evaluate the results on the KITTI online benchmark. 

\paragraph{Implementation Details.}
We implement AnyFlow in PyTorch~\cite{pytorch}. During training, we use the AdamW~\cite{adamw} optimizer with one-cycle learning rate policy and gradient clipping~\cite{raft}. We use 12 flow updates for training, 32 for Sintel evaluation and 24 for KITTI evaluation. For the implicit flow upsampler, we set $n$ as $4$. We initialize $r_0=4$ for dynamic lookup and $r_0=6$ for region encoding, respectively. During the multi-scale warping, we estimate $\mathbf{f}_{1/2}$ by the flow upsampler as mentioned in Sec.~\ref{sec:HR}, and we compute $\mathbf{f}_{1/4}$ by applying average pooling to $\mathbf{f}_{1/2}$. Further details are described in the supplementary material.

\subsection{Comparison with State-of-the-arts}
\paragraph{Cross-dataset generalization.}
In the left half of Table~\ref{table:comparison}, we evaluate the results on the training set of Sintel~\cite{sintel} and KITTI~\cite{kitti} to compare how each method can generalize on the target datasets. All methods are trained on the FlyingChairs~\cite{flownet} and FlyingThings~\cite{things}.

On the Sintel clean, AnyFlow ({\it R.E.}) achieves 1.10 EPE, reducing the error of the baseline, RAFT, by 23\%. We achieve the largest improvement among the recent RAFT-based methods~\cite{craft, gma, sun2022skflow, separableflow, flow1d, luo2022learning}, even though we use only 0.1M more parameters than RAFT. 
GMFlow~\cite{xu2022gmflow} and FlowFormer~\cite{huang2022flowformer} adopt Transformer~\cite{liu2021swin} for global matching to capture large displacements and achieve the best results.
Instead, AnyFlow captures large motions by dynamic control of correlation lookup and achieves competitive performance. With warm-start strategy~\cite{raft}, AnyFlow ({\it R.E.}) achieves the second best result, 1.05 EPE.
On the KITTI, AnyFlow outperforms every previous method by large margin in both metrics, which improves the baseline by more than 25\%. As the KITTI dataset contains relatively smaller motions compared to the Sintel, the proposed way of utilizing high-resolution features is effective to capture them. In addition, AnyFlow updates lookup range depending on input frames and generalizes well on the real-world dataset, KITTI, even though the training is done only on the synthetic datasets.

\paragraph{Test set benchmark.}
In the right half of Table~\ref{table:comparison}, we evaluate the results on the test set of Sintel and KITTI datasets. AnyFlow ({\it dynamic}) achieves the second best result (1.23 EPE) on Sintel clean with large improvements over RAFT. 
Since our modifications to the baseline are orthogonal to others, AnyFlow is compatible with previous works. We combine GMA~\cite{gma} with AnyFlow (AnyFlow + GMA) to demonstrate AnyFlow can further benefit from them. Several previous methods~\cite{craft, sun2022skflow, huang2022flowformer} also adopt GMA to resolve occlusion. AnyFlow achieves 1.21 EPE on Sintel clean after leveraging it.  

On the KITTI dataset, AnyFlow ({\it dynamic}) achieves 4.41 F1-all and ranks second among the previous studies. Note that we do not compare with the methods~\cite{jeong2022imposing, sun2022disentangling} that have no difference in architecture from our baseline, other than training strategy and datasets. Even though DIP~\cite{zheng2022dip} shows better result on this metric, we outperform it on all the other metrics. This result further demonstrates that AnyFlow generalizes well in real-world scenarios. 

Overall, AnyFlow performs well in general and is the best choice when considering diverse scenarios: cross-dataset generalization, test set results and number of parameters. Even though FlowFormer~\cite{huang2022flowformer} performs best on Sintel, AnyFlow is better on the real-world dataset, KITTI, with three times less parameters (18.2M vs 5.4M). Also, FlowFormer requires ImageNet~\cite{imagenet_cvpr09} pretraining to achieve good accuracy. We also report the results of FlowFormer (small) that uses less parameters without ImageNet training. We show better results on cross-dataset generalization. 

\subsection{Ablation Study}

In Table~\ref{table:ablation}, we conduct ablation studies to examine the impact of each component on the overall performance. The model is trained on synthetic datasets (C+T) and evaluated on the training sets of Sintel and KITTI.

In the upper section, we evaluate various strategies for multi-scale feature warping. MS {\it bilin.} employs bilinear upsampling on $\mathbf{f}_i$ to derive $\mathbf{f}_i^{1/2}$ and $\mathbf{f}_i^{1/4}$ for warping, while MS calculates intermediate flows using the implicit upsampler, as described in Sec.~\ref{sec:HR}. 

MS {\it bilin.} exhibits significant improvements over w/o MS across all metrics, and MS further enhances the results. The findings reveal that employing high-resolution features effectively increases accuracy, while the proposed methods contribute additional synergistic effects.

Essentially, the accumulated flow $\mathbf{f}_i$ is not directly supervised by ground-truth during training, but the flow after convex upsampling, i.e., $\mathbf{f}_i^1$, contributes to the loss computation. As a result, neither the accumulated flow nor the bilinearly upsampled one provides precise estimations.

\begin{figure}[h]
    \centering
    \includegraphics[width=0.975\columnwidth]{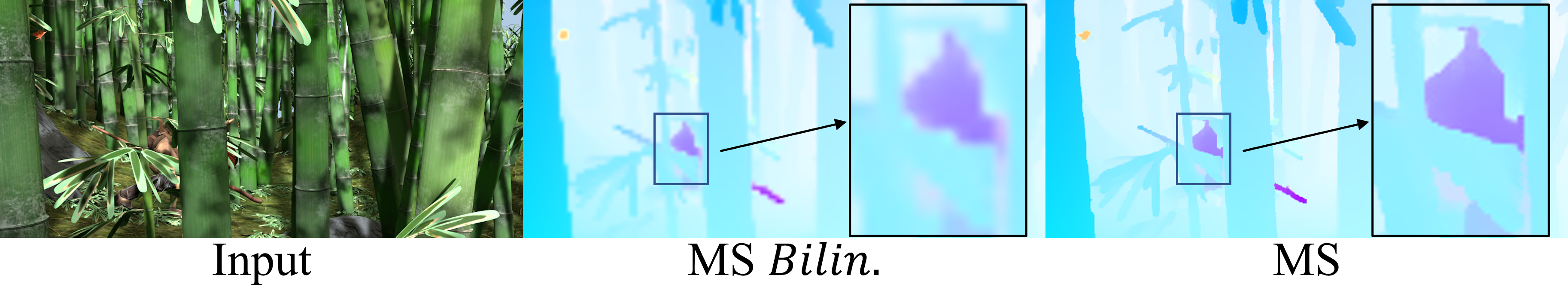}
    \caption{Comparison of upsampling strategies for multi-scale feature warping. We produce $\mathbf{f}^{1/2}_i$ by bilinear interpolation (MS {\it Bilin.}) and the proposed implicit upsampler (MS), respectively. 
    }
    \label{fig:hr_ablation}
\end{figure}

In contrast, our proposed implicit upsampler performs convex upsampling for any size, including $s \in \{1/2, 1/4\}$. This approach allows for accurate estimation of intermediate flows while preserving high-frequency details, making it more appropriate for warping.  We compare the visualizations in Fig.~\ref{fig:hr_ablation}.

The middle section reports the effect of the neural implicit flow upsampler (NIFU). We use the original RAFT upsampler for w/o NIFU. Note that we adopt MS {\it bilin.} in this ablation, since the complete MS warping leverages the implicit upsampler. As shown in the table, we obtain large improvements especially on the KITTI. Owing to the continuous representation, we can estimate precise boundaries and adopt multi-scale training to train our network under diverse motion ranges. Since it consists of MLP, it produces better results with less parameters. Not only improving the quantitative performance, the implicit upsampler also produces synergistic effect with the MS warping strategy and enables flow generation in arbitrary scales. Further ablations and the effects of multi-scale training can be found in our supplementary material.

\begin{table}[t]
\centering
\begin{tabular}{@{}c c c c c c @{}}
\toprule
\multirow{2}{*}{Method} & \multicolumn{2}{c}{Sintel} & \multicolumn{2}{c}{KITTI } & Param                                    \\
\cmidrule(lr){2-3} \cmidrule(lr){4-5} 
      &  Clean & Final & F1-epe & F1-all                                  \\
\midrule

w/o MS    & 1.29 & 2.72 & 4.02 & 13.91 & 5.04                         \\
MS {\it bilin.} & 1.16 & 2.55 & 3.81 & 12.82 & 5.39                   \\
MS & {\bf 1.10} & {\bf 2.52} & {\bf 3.76} & {\bf 12.44} & 5.39         \\
\midrule
 w/o NIFU  &  1.24	& {\bf 2.53} &	4.47 &	14.39  & 5.62             \\
 NIFU & 	{\bf 1.16} &	2.55 & {\bf 3.81} & {\bf 12.82} & 5.39    \\
\midrule
Fixed & 1.21 & 2.68 & 4.41 & 14.28 & 5.38                             \\
Dynamic & 1.17 & 2.58 & 3.95 & 13.01 & 5.39                           \\
Region E. & {\bf 1.10} & {\bf 2.52} & {\bf 3.76} & {\bf 12.44} & 5.39 \\
\bottomrule
\end{tabular}
\caption{{\bf Ablation experiments.} The models are trained on the synthetic datasets (C+T), and evaluated on Sintel and KITTI.}
\label{table:ablation}
\end{table}

In the last section, we compare the lookup strategies. Since the dynamic lookup controls the lookup ranges depending on the input images, it can focus on more optimal candidates to compute accurate flow. It shows large improvements especially on KITTI, which demonstrates its effectiveness for real-world generalization. The region encoding (Region E.) further improves the result in all metrics. As shown in Table~\ref{table:comparison}, on the other hand, AnyFlow ({\it dynamic}) shows better results on the test set.

\begin{figure*}[t!]
    \centering
    \includegraphics[width=0.975\linewidth]{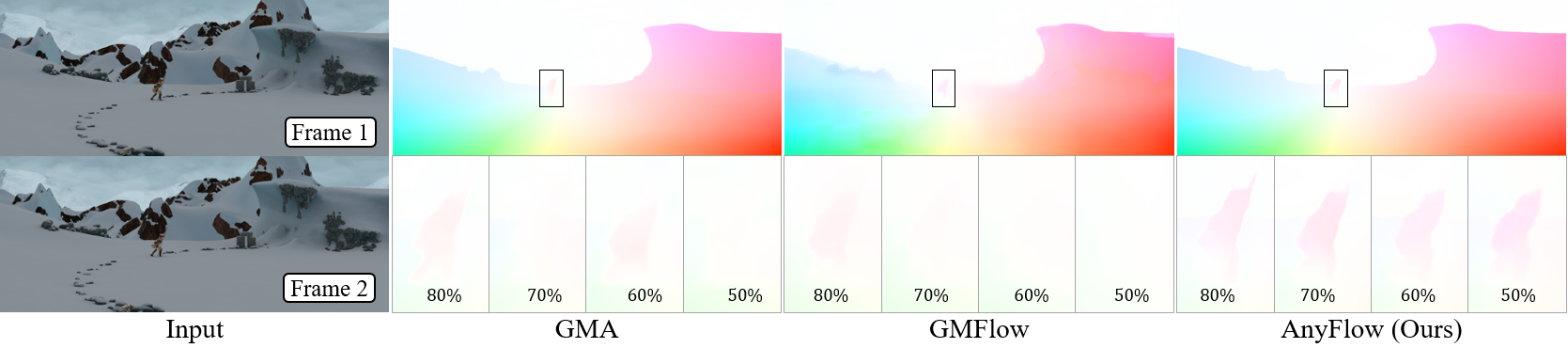}
    \caption{Qualitative comparison on Sintel test images against GMA~\cite{gma} and GMFlow~\cite{xu2022gmflow}. The top row is from input frames without resize and the bottom row showcases the close-up for the corresponding results by different down-scale ratios, indicated by the numbers.}
    \label{fig:qualitative}
\end{figure*}

\subsection{Arbitrary Scale Optical Flow Estimation}

\paragraph{Robustness on downsampled images.} 
In Table~\ref{table:robust}, we compare robustness to downsampling with the recent methods. We feed input to the networks as downsampled images by multiple scales from $50\%$ to $90\%$. The produced outputs are restored to the original shape and used to compute EPE between ground-truth. All models are trained on the synthetic datasets (C + T) and evaluated on Sintel and KITTI. 

On the Sintel dataset in Table~\ref{table:robust} (a), AnyFlow shows only $1\%$ of degradation up to $80 \%$ of downsampling. Up to $70\%$ of downsampling, which is a typical range of downsampling ratio in practice, the EPEs increase by $0.21$ for RAFT, $0.32$ for GMA, and $0.28$ for GMFlow.  AnyFlow shows only 0.1 error increase, which demonstrates its robustness in practical scenarios where downsampling is used. 
In the more extreme cases, i.e., $60 \%$ and $50\%$ of downsampling, other methods severely suffer from accuracy loss. 
In both datasets, AnyFlow achieves competitive accuracy even with $50\%$ downsampled images, which is still better than the original results ($100\%$) of the baseline, RAFT.

In Fig.~\ref{fig:intro} and Fig.~\ref{fig:qualitative}, we qualitatively compare the predictions. Since the metric is averaged over every pixel, it does not reflect well how each method can detect small objects under downsampled images. As shown in the figures, recent methods~\cite{raft, gma, xu2022gmflow} lose ability to capture small objects since they are dealing with low-resolution intermediate features. On the other hand, AnyFlow restores more clear boundary and captures small objects well with images of any sizes. Thanks to the proposed multi-scale warping and dynamic control of the correlation range, AnyFlow performs more robust estimation on diverse resolutions.
\begin{table}[t]
\small
  \begin{subtable}[c]{\columnwidth}
  \centering
  \begin{tabular}{@{}c c c c c c c @{}}
    \toprule
    Method &  100\% & 90\% & 80\% & 70\% & 60\% & 50\%  \\
    \midrule
    RAFT~\cite{raft} & 1.47 & 1.51 & 1.60 & 1.68 & 1.81 & 1.92 \\
    GMA~\cite{gma} & 1.31 & 1.43 & 1.45 & 1.63 & 1.71 & 1.92\\
    GMFlow~\cite{xu2022gmflow} & {\bf 1.08} & 1.18 & 1.17 & 1.36 & 1.71 & 2.35\\
    AnyFlow & 1.10 & {\bf 1.12} &{\bf 1.11} & {\bf 1.20} & {\bf 1.31} & {\bf 1.41} \\
    \bottomrule
  \end{tabular}
      \caption{EPE on Sintel (clean)}

  \end{subtable}
  \vspace{1mm}

    \begin{subtable}[c]{\columnwidth}
  \centering
  \small
  \begin{tabular}{@{}c c c c c c c @{}}
    \toprule
    Method &  100\% & 90\% & 80\% & 70\% & 60\% & 50\%  \\
    \midrule
    RAFT~\cite{raft} & 5.04 & 5.22 & 5.39 & 6.17 & 6.61 & 7.50 \\
    GMA~\cite{gma} & 4.48 & 4.60 & 4.79 & 5.21 & 5.67 & 6.25 \\
    GMFlow~\cite{xu2022gmflow} & 7.49 & 7.81 & 8.76 & 9.63 & 9.17 & 10.41 \\
    AnyFlow &  {\bf 3.76} & {\bf 3.80} & {\bf 3.92} & {\bf 3.97} & {\bf 4.28} & {\bf 4.88}
  \\
    \bottomrule
  \end{tabular}
    \caption{EPE on KITTI}
  \end{subtable}
  \caption{Quantitative comparisons of performance with downsampled images on Sintel and KITTI datasets.}
  \label{table:robust}
\end{table}
\paragraph{Generating high-resolution optical flow.}

\begin{figure}[t!]
    \centering
    \includegraphics[width=0.975\columnwidth]{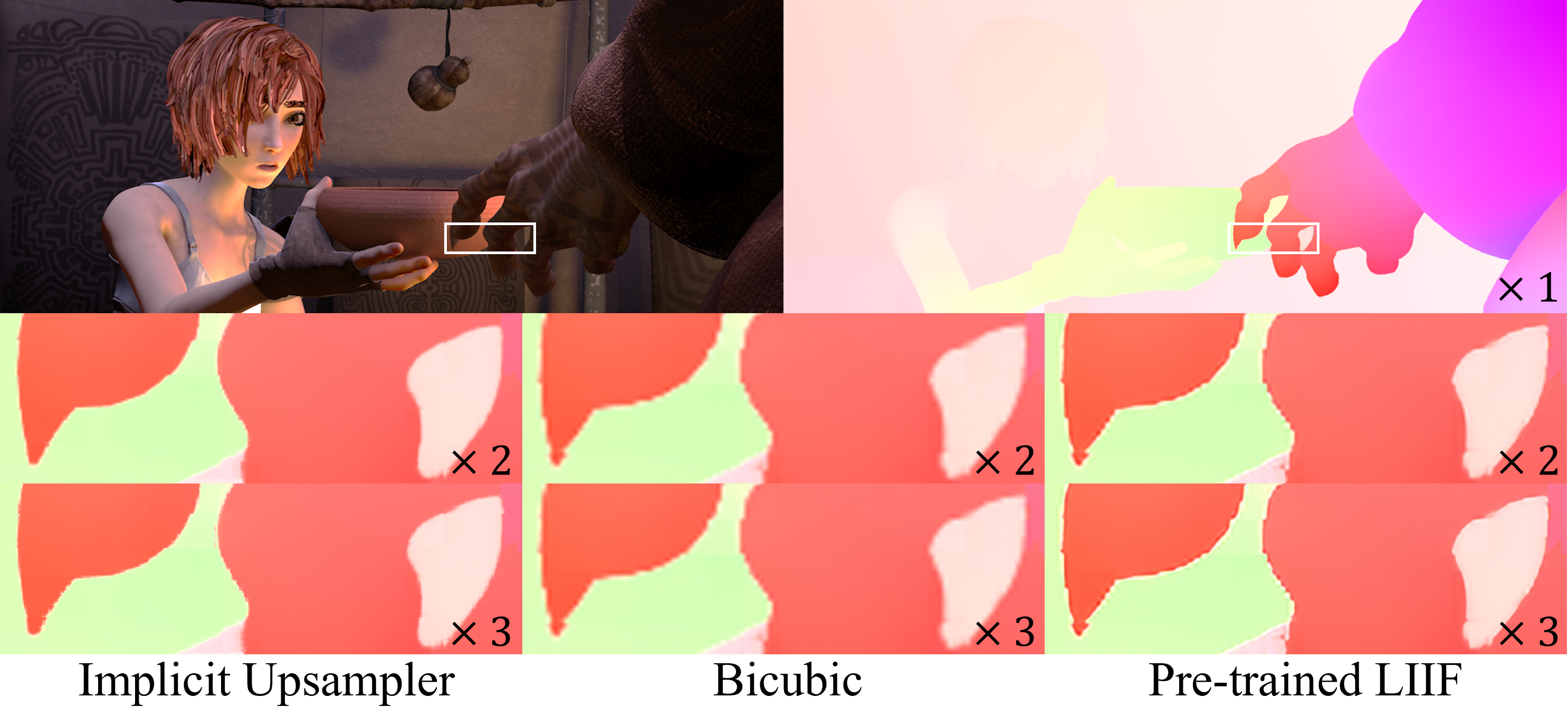}
    \caption{We evaluate our method (left) on flow upsampling against bicubic (middle) and results by using pre-trained LIIF~\cite{liif} (right). Compared to both methods, our results produce sharper object boundaries without introducing artifacts.}
    \label{fig:upsample}
\end{figure}

In Fig.~\ref{fig:upsample}, we compare the results of upsampling at higher resolutions ($\times 2$, $\times 3$) using different methods. To compare with a super-resolution method, we first convert the flow output to a 3-channel RGB and feed it into the pre-trained LIIF network~\cite{liif}. For both interpolation and super-resolution, we initially estimate the output at the original scale, $\mathbf{f}_i^1$, and then upsample it. On the other hand, the implicit upsampler produces high-resolution flow, $\mathbf{f}_i^2$ and $\mathbf{f}_i^3$, directly from the accumulated flow, $\mathbf{f}_i$. As illustrated in the figure, interpolation and LIIF introduce blocking artifacts because they use low-resolution flow ($\times 1$) as input, amplifying noise and artifacts as resolution increases. In contrast, we train the network end-to-end, allowing the implicit upsampler to directly generate high-resolution flow from the latent space. This results in clearer boundaries and fewer artifacts.

\section{Conclusion}

We have proposed AnyFlow, a new method that models optical flow as a continuous coordinate-based representation, enabling the generation of outputs at arbitrary scales. With the novel warping module and dynamic lookup strategy, AnyFlow exhibits robust performance across various motion types and input resolutions. In particular, it substantially enhances estimation quality for low-resolution images and small objects while preserving intricate details. To our knowledge, AnyFlow is the first approach designed to provide robust estimations for low-resolution inputs, which extends its applicability to portable devices.

{\small
\bibliographystyle{ieee_fullname}
\bibliography{main}
}

\end{document}